\documentclass[review,11pt]{ReportTemplate}
\usepackage{bm}

\usepackage{hyperref}
\usepackage{url}

\usepackage{algorithm}
\usepackage[noend]{algorithmic}

\usepackage{color}
\usepackage{subfigure}
\usepackage{amsmath}
\usepackage{amssymb}
\usepackage{mathrsfs}
\usepackage{color}
\usepackage{algorithmic}
\usepackage{epsfig}
\usepackage{epsf}
\usepackage{graphics}

\usepackage{multicol}
\usepackage{multirow}
\usepackage{url}

\usepackage{graphicx}
\usepackage{subfigure}

\def \x {{\bf x}}
\def \z {{\bf z}}
\def \y {{\bf y}}
\def \w {{\bf w}}
\def \e {{\bf e}}
\def \thta {{\bf \theta}}
\def \c {{\bf c}}
\def \N {{ \textit{N}}}

\def \z {{\bf z}}

\def\0{{\bf 0}}
\def\1{{\bf 1}}

\begin{document}
\begin{frontmatter}
\title{Multi-Label Active Learning from Crowds}

\author{Shao-Yuan Li}
\author{Yuan Jiang}
\author{Zhi-Hua Zhou\corref{cor1}}
\address{National Key Laboratory for Novel Software Technology\\
Nanjing University, Nanjing 210093, China} \cortext[cor1]{\small Corresponding author.
Email: zhouzh@nju.edu.cn}

\begin{abstract}
Multi-label active learning is a hot topic in reducing the label cost by optimally choosing the most valuable instance to query its label from an oracle. In this paper, we consider the pool-based multi-label active learning under the crowdsourcing setting, where during the active query process, instead of resorting to a high cost oracle for the ground-truth, multiple low cost imperfect annotators with various expertise are available for labeling. To deal with this problem, we propose the MAC (Multi-label Active learning from Crowds) approach which incorporate the local influence of label correlations to build a probabilistic model over the multi-label classifier and annotators. Based on this model, we can estimate the labels for instances as well as the expertise of each annotator. Then we propose the instance selection and annotator selection criteria that consider the uncertainty/diversity of instances and the reliability of annotators, such that the most reliable annotator will be queried for the most valuable instances. Experimental results demonstrate the effectiveness of the proposed approach.

\end{abstract}

\end{frontmatter}

\section{Introduction}\label{sec:intro}
  Multi-label learning has received significant attention to deal with examples that are associated with multiple classes simultaneously~\cite{MultiLabelReviewZhou}. It is worth noting that it is usually quite expensive to obtain the labels for training, as every possible label has to be checked whether it is proper for the instance. Thus multi-label active learning, which reduces the labeling cost by actively selecting the most valuable instances to query becomes a hot topic~\cite{MultLabelActiveijcai13,MultLabelActiveicdm13}.

  Most multi-label active learning algorithms rely on the domain expert, or an oracle, to provide the ground-truth label for each query. While the labeling cost of domain experts is high, by employing multiple low cost non-expert labelers, crowdsoucing provides a low cost way to get annotations. A number of studies exploiting the wisdom of crowds have arisen recently~\cite{CrowdSourcingSnowNLP08,Raykar10,Welinder10,DennyZhouICML14}, mainly focusing on single-label tasks where each instance is associated with only one label. Exploiting the wisdom of crowds for multi-label data has barely been touched to the best of our knowledge.

  In this paper, we consider the problem of exploiting the wisdom of crowds for multi-label tasks. While multiple annotators may be available, in real-world applications, the labeling cost would still be high for multi-label objects considering that every label needs to be checked on each instance by several annotators, whereas the annotation budget is often limited. Thus we define our problem from the active learning perspective, where instances are actively selected to query the supervised information, but rather than a high cost oracle providing the ground-truth, multiple imperfect annotators are available for labeling. The goal of our work is to learn an effective multi-label classifier with as less annotation cost as possible.

  By decomposing the multi-label task into a series of independent binary classification problems, Y. Yan's work~\cite{CrowdactiveYan11} can be applied to each label independently, neglecting the fact that the information of one label may be helpful for learning another related label, especially when there are insufficient training data for every label. To handle the problem of multi-label active learning from crowds, we propose the MAC (Multi-label Active learning from Crowds) approach, which incorporates the local influence of label correlations to build a probabilistic model for multi-label classifier and annotators. Based on this model, we can estimate not only the labels of instances but also the expertise of annotators. Then we propose our instance selection criterion which considers both uncertainty and diversity of instances, and the most reliable annotator on the most valuable instance is queried.

  In the following we start with a brief review of some related work. Then, we propose our MAC approach and report
  the experimental results. Finally, we conclude the paper .

  \section{Related Work}
% multi-label active learning
 \textbf{Multi-label Active Learning}  Multi-label learning deals with examples that are associated with multiple labels simultaneously; it has received significant attention during the past decades~\cite{MultiLabelReviewZhou} and achieved successful application in various tasks such as image annotation~\cite{MultiLabelImage}, gene function classification~\cite{MultiLabelGene} and text categorization~\cite{MultiLabelText}. To collect labels for multi-label training, each of the multiple labels should be checked whether it is proper for an instance; the labeling cost is quite expensive. Thus multi-label active learning, which reduces the labeling cost by actively selecting the most valuable instance to query from an oracle, has attracted great attention. Existing multi-label active learning research mainly focus on designing the criterion for instance selection, and can be roughly categorized into three categories: 1) uncertainty sampling~\cite{MultLabelActiveSigh08} which selects the instance that the classifier is most uncertain about ; 2) expected loss reduction~\cite{MultLabelActiveacml11} which queries instance that minimizes the expected loss of the classifier; and 3) combining multiple criteria~\cite{MultLabelActiveijcai13,MultLabelActiveicdm13} to select instance. Given the selected instance, most work~\cite{MultLabelActiveacml11,MultLabelActiveijcai13} query all the labels for the instance, which may lead to information redundancy and wasting of oracle's effort, since labels are often correlated in multi-label learning.% To avoid the information redundancy, Huang etal.~\cite{MultLabelActiveicdm13} considered querying a instance-label pair, i.e.,  a specific label's relevance to the selected instance.

% crowdsourcing learning
  \textbf{Crowdsourcing} With the advent of crowdsourcing platforms such as Amazon Mechanical Turk (AMT), crowdsourcing which makes use of crowdsourced annotations from multiple imperfect labelers is widely used for tasks including sentiment classification~\cite{CrowdSourcingSnowNLP08}, medical diagnosis~\cite{ Raykar10}, image tagging~\cite{Welinder10} and webpage categorization~\cite{DennyZhouICML14}. As annotators may have different expertise on different tasks, the common wisdom is to distribute tasks to multiple annotators and then estimate the correct labels via some aggregation schemes. Simply taking the majority voting without considering the annotators ability variance may lead to poor impact on subsequent learning. Hence, in order to solve the above problem of annotators, a number of methods assessing abilities of annotators have been proposed. The annotator's expertise on specific task are usually modeled by either measures with explicit explanation like accuracy~\cite{Whitehill09,KOS11} and confusion matrix~\cite{Raykar10,DennyZhouICML14}, or complex high multidimensional vectors~\cite{Welinder10}.
 % active crowdsourcing
  A few work trying to get good learning performance with low crowds cost are proposed, for example, ~\cite{CrowdactiveYan11,Crowdactivemoe14} consider the cost by adaptively assigning the best annotator to specific task under the active learning framework and ~\cite{CrowdcostAMV14} propose an approach to approximate the performance of majority voting with less annotations.

  While current work on crowdsourcing mainly focus on single-label tasks, to our best knowledge, using crowds in a cost economic way has not been studied in multi-label learning. We are presenting possibly the first approach to exploiting the wisdom of crowds efficiently in  multi-label learning.

  \section{MAC: Multi-Label Active learning from Crowds}
 We consider the pool-based multi-label active learning from crowds, where a set of labeled annotated data of $N_l$ examples $D_l=\{({\x}_1,\z_{1},\{{\y_{1j}}\}),\ldots,({\x}_{N_l},\z_{N_l},\{{\y_{{N_l}j}}\})\}$  and an unlabeled set of $N_u$ examples $D_u =\{ {\x}_{N_l+1},\ldots, {\x}_{N_l+N_u}\}$  are available for training. Each instance ${\x}_i$ is a $d$-dimensional feature vector. Given an instance ${\x}_i\in D_l$, ${\z}_i \in \{+1, -1\}^{L\times 1}$ is its true label assignment, where $L$ is the number of labels and  $\z_{i}^{l}= +1(-1)$ indicates that the $l$-th label is tagged as positive(negative) for $\x_i$; and ${\y_{ij}}\in \{+1, -1, 0\}^{L\times 1}$ is the label assignment given by annotator $j$, where $\y_{ij}^{l}=0$ means that the annotator $j$ gives no annotation for $\x_i$ on the $l$-th label , and $\y_{ij}^{l}= +1(-1)$ means annotator $j$ tags the $l$-th label as positive(negative) for $\x_i$. We assume a set of $M$ annotators $\{{\w}_j\}_j$ where $j \in \{1,2,\ldots,M \}$ are available. We do not require each instance to be annotated by all annotators on every label, and the annotator set labeling instance $\x_i$ on label $l$ is denoted as $M_i^l$, i.e., $M_i^l=\{j|\y_{ij}^{l}=1(-1)\}$. During the active learning process, rather than an oracle being available to provide the groundtruth labels $\z$, multiple imperfect annotators with various expertise are available for annotating. The target of our work is to actively select the most reliable annotator for the most valuable instance such that we can learn an effective multi-label classifier with as less labeling cost as possible.

 Compared with traditional multi-label active learning, we can see that the key challenges in the active learning from crowds lie in that: 1) the annotators reliability can be various for specific instance, which requires the careful selection of the annotator with the best expertise for a query; 2) the labels provided by selected annotator for specific queried instance can be noisy, which requires a further aggregation step to get a better estimation of the groundtruth label. To solve the above challenges, we use a probabilistic model which can simultaneously inference the annotators expertise and groundtruth label at the same time. Furthermore, the local influence of neighborhoods' label correlations are incorporated in this probabilistic model to help the multi-label classifier  learning. After we obtain the estimation of annotators expertise and instances labels, we can design some criterion for instance and annotator selection to get the most reliable annotation for the most valuable instance. In the following subsections, we first propose our probabilistic multi-label crowdsourcing model and then the active selection strategy.

\subsection{Probabilistic Multi-Label Crowdsourcing Model}
 We first introduce our probabilistic model on one single label to describe the classifier and annotators, and then encode the local influence of neighborhoods' label correlations in this model to deal with multi-label crowdsourcing.
\begin{figure}[htbp]
  \centering
  \subfigure[Single Label Model]{
    \label{fig:singlegf} %% label for first subfigure
    \includegraphics[width=2.6in]{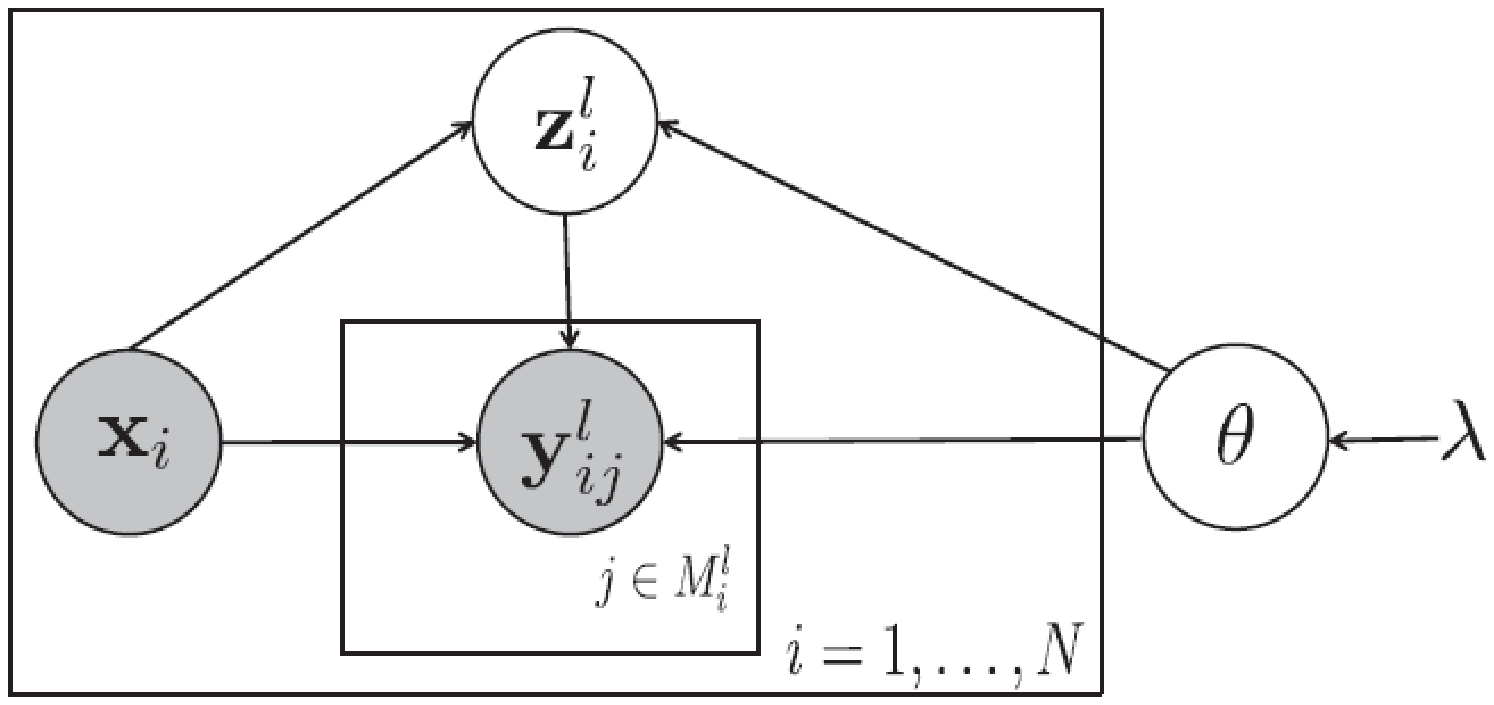}}
    \hspace{0.1in}
  \subfigure[Our Multi-Label Model]{
    \label{fig:multigf} %% label for second subfigure
   \includegraphics[width=2.6in]{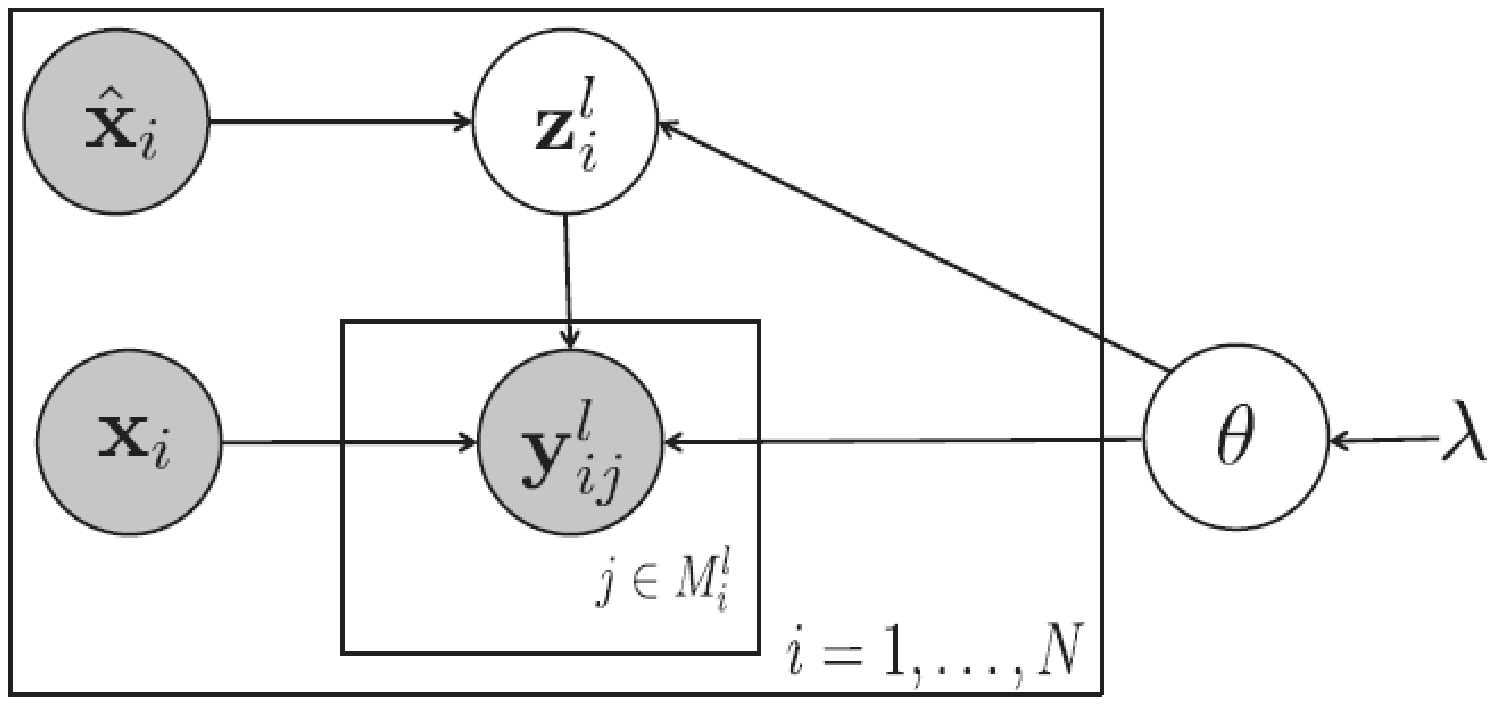}}

  \caption{(a) The single label graphical model for instances $\{\x_i\}$, annotations $\{\y_{ij}^{l}\}$ and unknown groundruth $\{\z_{i}^{l}\}$; (b) Our multi-label graphical model for instances $\{\x_i\}$, $\{\hat{\x}_i\}$, annotations $\{\y_{ij}^{l}\}$ and unknown groundruth $\{\z_{i}^{l}\}$ on label $l$.}
  \label{fig:graphicalmodel} %% label for entire figure
\end{figure}

 Figure~\ref{fig:singlegf} illustrate the probabilistic graphical model on label $l$ over the instances $\{\x_i\}$, the observed annotations given by annotators $\{\y_{ij}^{l}\}$, and the unobserved groundruth labels $\{\z_{i}^{l}\}$. We assume that the annotations given by the annotators depend both on what input they observe and what task they are expected to finish, i.e., the input instance $\x$ and the target groundtruh label $\z$. The groundtruth label of the instance is believed to reflect specific property of the instances, for example, $\z_i=1/-1$ may denote the presence/absence of the tree in a picture. Using $\thta$ to denote the parameters involved in this model and $p_r(\thta)$ its prior probability distribution, the probabilistic graphical model can be represented using the following joint probability
 distribution:
 \begin{eqnarray}\label{eq:jointProb}
 P( \{\y_{ij}^{l}\}_{ij},\{\z_{i}^{l}\}_i| \{{\x_i}\}_i, \thta )= \prod\limits_i{ p( \z_{i}^{l}|{\x_i},\thta)} \prod\limits_{j\in {M_i^l}} {p(\y_{ij}^{l}|\z_{i}^{l},{\x_i}, \thta)} p_r(\thta).
 \end{eqnarray}

 For each annotator $j$,  we define an expertise level variable $\e_{ij}^{l}$ to build the relationship between its annotation $\y_{ij}^{l}$ and the input $({\x_i,\z_{i}^{l}})$, which is mathematically define as the probability that annotator $j$ provides the groundtruth as its annotation for instance $i$. Here we exploit a \emph{logistic sigmoid} acting on some function $f_{j,\thta}^{l}$ over the instance ${\x_i}$, that is,
 \begin{eqnarray}\label{eq:expertise}
  \e_{ij}^{l}=p(\y_{ij}^{l}= \z_{i}^{l}|{\x_i}, \thta )= \sigma(f_{j,\thta}^{l}({\x_i}) ),
  \end{eqnarray}
  where the logistic sigmoid function is defined as $\sigma(z)=1/(1+exp(-z))$. Using $I(\cdot)$ to denote the indicator function, then the distribution $p(\y_{ij}^{l}|\z_{i}^{l},{\x_i},\thta)$ can be formulated as the following Bernoulli distribution,
  \begin{eqnarray}\label{eq:annotProb}
   p(\y_{ij}^{l}|\z_{i}^{l},{\x_i}, \thta)=(1-\e_{ij}^{l})^{[1-I(\y_{ij}^{l}=\z_{i}^{l})]} (\e_{ij}^{l})^ {[I(\y_{ij}^{l}=\z_{i}^{l})]}.
   \end{eqnarray}

  Since we are dealing with classification, we use a \emph{logistic sigmoid} acting on some function $f_{0,\thta}^{l}$ over instance ${\x_i}$ to model the Bernoulli distribution  $p( \z_{i}^{l}|{\x_i},\thta)$ for $\z$, that is,
  \begin{eqnarray}\label{eq:classifier}
  p( \z_{i}^{l}=1|{\x_i},\thta)= \sigma(f_{0,\thta}^{l}({\x_i})) & and & p( \z_{i}^{l}=-1|{\x_i},\thta)= \sigma(-f_{0,\thta}^{l}({\x_i})).
  \end{eqnarray}
  While any function can be used to implement $f_{0,\thta}^{l}$ and $f_{j,\thta}^{l}$, for ease of exposition, we consider the linear discriminating functions, i.e., $f_{0,\thta}^{l}(\x_i)=(\w_{0}^{l})^{'}{\x_i} $ and $f_{j,\thta}^{l}(\x_i)=(\w_{j}^{l})^{'}{\x_i}$.
 Given this, the parameters becomes the classifier/annotator parameters $\thta =\{\w_{0}^{l}, \{\w_{j}^{l}\} \}$. To overcome overfitting, we introduce a zero-mean $\lambda$-variance Gaussian prior for $\{\w_{0}^{l}\}$ and $\{\w_{j}^{l}\}$, i.e., :
   \begin{eqnarray}\label{eq:paraPrior}
  p_r(\thta) = p(\w_{0}^{l}|\lambda) \prod\limits_{j\in M_i^l}{p(\w_{j}^{l}|\lambda)} &  where &
  p(\w_{0}^{l}|\lambda)= p(\w_{j}^{l}|\lambda)=\mathcal{N}(\textbf{0},\lambda^{-1}\textbf{{I}})
  \end{eqnarray} We can estimate the parameters by exploiting the maximum likelihood criterion, to solve which a standard Expectation-Maximization (EM) approach~\cite{EM77} can be used with missing variables $\{\z_{i}^{l}\}$ and observed variables $\{\y_{ij}^{l}\}_{ij}$.

 \textbf{E-step:} Given current estimation of the parameters $\thta =\{\w_{0}^{l}, \{\w_{j}^{l}\} \}$ from last M step, the posterior probability of ground truth label $\{\z_{i}^{l}\}$ is computed:
 \begin{eqnarray}\label{eq:Estep}
   & p( \z_{i}^{l}) & = p( \z_{i}^{l}|{\x_i},\{\y_{ij}^{l}\}_{ij}, \w_{0}^{l}, \{\w_{j}^{l}\}) \nonumber\\
   & &  \propto  p( \z_{i}^{l}|{\x_i},\w_{0}^{l}) \prod\limits_{j\in {M_i^l}} {p(\y_{ij}^{l}|\z_{i}^{l},{\x_i},\w_{j}^{l} )}.
 \end{eqnarray} %The predicted label ${\hat{\z}_{i}}^{l}$ of ${\x_i}$ is set as 1 if $p( {\z_{i}}^{l})>0.5$ and $-1$ otherwise.
Substituting Eq.~\ref{eq:annotProb} and  Eq.~\ref{eq:classifier} into Eq.~\ref{eq:Estep}, E-step reduces to:
  \begin{eqnarray}\label{eq:EstepImplement}
  &p( \z_{i}^{l}=1) &  \propto \;  \sigma({\w_{0}^{l}}^{'}{\x_i}) \prod\limits_{\y_{ij}^{l}=1}\sigma({\w_{j}^{l}}^{'}{\x_i})  \prod\limits_{\y_{ij}^{l}=-1}\sigma(-{\w_{j}^{l}}^{'}{\x_i})  \\
  &p( \z_{i}^{l}=-1) &  \propto \;  \sigma(-{\w_{0}^{l}}^{'}{\x_i}) \prod\limits_{\y_{ij}^{l}=-1}\sigma({\w_{j}^{l}}^{'}{\x_i})  \prod\limits_{\y_{ij}^{l}=1}\sigma(-{\w_{j}^{l}}^{'}{\x_i})
  \end{eqnarray}

 \textbf{M-step:} To estimate the parameters $\thta =\{\w_{0}^{l}, \{\w_{j}^{l}\} \}$, we maximize the expectation of the joint log-likelihood of $(\{\y_{ij}^{l}\}_{ij},\{\z_{i}^{l}\}_i)$ over parameters $\thta =\{\w_{0}^{l}, \{\w_{j}^{l}\} \}$, with respect to the posterior probabilities of  $\{\z_{i}^{l}\}$ computed by last E step:
 \begin{eqnarray}
 \thta = \arg\max_\thta Q(\thta)  \nonumber,
 \end{eqnarray} where $Q(\thta)$ is
  \begin{eqnarray} \label{eq:Q}
  &  Q(\theta) & =\;\;  E_{\z}[\ln P( \{\y_{ij}^{l}\}_{ij},\{\z_{i}^{l}\}_i| \{{\x_i}\}_i,\w_{0}^{l}, \{\w_{j}^{l}\} ) p_r(\thta) ] \nonumber\\
  &&= \;\;   E_{\z}[\ln  \prod\limits_i{ p( \z_{i}^{l}|{\x_i},\w_{0}^{l})p(\w_{0}^{l}|\lambda)}\; \prod\limits_{j  } {p(\y_{ij}^{l}|\z_{i}^{l},{\x_i}, \{\w_{j}^{l}\})}p(\w_{j}^{l}|\lambda)  ] \nonumber\\
  &&= \;\;\sum\limits_i E_{\z}[\ln  p( \z_{i}^{l}|{\x_i},\w_{0}^{l})]  + \sum\limits_{ij} E_{\z}[\ln p(\y_{ij}^{l}|\z_{i}^{l},{\x_i},\w_{j}^{l}) ]   \nonumber\\
  &&\;\;\;\;\;\;+\frac{\lambda}{2}\|\w_{0}^{l}\|^2+\frac{\lambda}{2}\sum\limits_j\|\w_{j}^{l}\|^2 .
  \end{eqnarray}
  Substituting Eq.~\ref{eq:annotProb} and  Eq.~\ref{eq:classifier} into Eq.~\ref{eq:Q}, $Q(\theta)$ reduces to:
  \begin{eqnarray} \label{eq:Qwol}
   & Q(\w_{0}^{l}) & = \sum\limits_i ({\w_{0}^{l}}^{'}{\x_i})p( \z_{i}^{l}=1) - \sum\limits_i [\ln (1+exp({\w_{0}^{l}}^{'}{\x_i}) ].\\
   & Q(\w_{j}^{l}) & = \sum\limits_i ({\w_{j}^{l}}^{'}{\x_i})p( \z_{i}^{l}=1)\cdot\y_{ij}^{l}  - \sum\limits_i \ln [1+exp({\w_{j}^{l}}^{'}{\x_i} \cdot\y_{ij}^{l})] .
  \end{eqnarray}
  The gradient of $Q$ with respect to parameters $\w_{0}^{l},\w_{j}^{l}$ is computed as:
  \begin{eqnarray} \label{eq:Qwolgradient}
   & \frac{\partial Q} {\partial{\w_{0}^{l}}} & = \sum\limits_i {\x_i}[p( \z_{i}^{l}=1) - \sigma({\w_{0}^{l}}^{'}{\x_i}) ].\\
   & \frac{\partial Q} {\partial{\w_{j}^{l}}}  & =\sum\limits_i {\x_i}[p( \z_{i}^{l}=1) - \sigma({\w_{j}^{l}}^{'}{\x_i}\cdot\y_{ij}^{l}) ]\cdot\y_{ij}^{l}.
  \end{eqnarray}
  To deal with multi-label data, one straightforward approach is to treat each label independently and applying the above model on each label separately. But by this, the label correlations are ignored, which however is widely verified to be helpful for multi-label learning. In this paper, we derive a local code to enhance the feature representation of instances, which reflects the local influence of its neighborhoods' label correlations. We use the simple idea that instances similar in the feature space should also be similar in the label space, and utilize the information from the label space of labeled instances to construct extra features. In detail, for each instance $\x$, its code vector $\c$ is constructed as the label average mean of its $k$ nearest neighbors in the initial labeled training set, i.e.,
 \begin{eqnarray} \label{eq:labelCorCode}
 \c = \frac{1}{k} \sum\limits_{\x_i \in \N(\x)} \z_i.
 \end{eqnarray}
 After the code vector is computed, the \emph{enhanced representation} of instance $\x$ becomes $\hat{\x}=[\x,\c]$. Then we learn the above probabilistic model on each label separately, but using different instance representations for classifier and annotators. For label $l$, the multi-label classifier parameter $\w_0$ is learned on the \emph{enhanced representation}, but for the annotator model parameters $\{\w_{j}^{l}\}$, the \emph{original representations} $\x$ are used, i.e., $\w_0$ is of dimension  $(d+L)$ and $\{\w_{j}^{l}\}$ are of dimension $d$. Utilizing the local influence of label correlations, our multi-label probabilistic crowsourcing model can be described by Figure~\ref{fig:multigf}. The reason we use different instance representations for the classifier and annotators is that, the \emph{enhanced representation} of instances are deduced from the initial labeled data, which are generated by the true multi-label data distribution, so it is expected that only for cases where the training data are consistent with the multi-label data distribution, the \emph{enhanced representation} are to be helpful for learning. While the groundtruth estimation aggregated from the multiple annotators should be trustable and  consistent with the multi-label data distribution, the annotations provided by each annotator can be of low quality and far away from the multi-label data distribution, thus the \emph{enhanced representation} generated from the labeled training data would be inconsistent with its labeling behavior.  Though here the way we compute the \emph{enhanced representation} is simple, we believe the idea of constructing new representations for the classifier using information from the label space would be a good direction for multi-label crowdsourcing.

\subsection{Active Selection}
 Based on the above model, we can learn an estimation of the multi-label classifier, the annotators expertise and the labels for instances from the crowds annotations. In traditional multi-label active learning, most work select the instance and query its supervised information on all labels, which may lead to information redundancy and wasting of annotation effort since labels are often correlated in multi-label learning. To avoid information redundancy, we exploit the idea of first choosing the most valuable instance according to its label estimation, and then selecting its most uncertain label to query the supervised information. Considering that the supervised information in our problem are provided by crowds whose reliability on different instances may be different, therefore, the selection of the most reliable annotator for the specific instance-label pair should also be carefully considered.

 \textbf{Instance-Label Pair Selection} We design our instance selection criterion by combining the uncertainty criterion with the diversity of queried annotations. LCI (instance Label Cardinality Inconsistency) is a commonly used instance uncertain measure in multi-label active learning and its combination with other measures to conduct instance selection has shown promising results in several works~\cite{MultLabelActiveijcai13,MultLabelActiveicdm13}. It is defined as the inconsistency between the number of predicted positive labels of instances and the average label cardinality on the labeled data,
 \begin{eqnarray}\label{eq:LCI}
 LCI(\x_i) = (\sum\limits_{l=1}^L I(\hat{\z}_{i}^{l}=1)- \frac{1}{N_l}\sum\limits_{j=1}^{N_l} \sum\limits_{l=1}^L I(\z_{j}^{l}=1)  )^2,  \nonumber
 \end{eqnarray} where $\hat{\z}_{i}^{l}$ is the label estimation of instance $\x_i$ on $l$. We extend LCI to incorporate the query diversity during the active process and define a new criterion
 \begin{eqnarray}\label{eq:instanceSel}
 i^\ast = \arg\max_i CI(\x_i) = \frac{|\sum\limits_{l=1}^L I(\hat{\z}_{i}^{l}=1)- \frac{1}{N_l}\sum\limits_{j=1}^{N_l} \sum\limits_{l=1}^L I(\z_{j}^{l}=1)|}{ max\{\xi, anno(\x_i)\} },
 \end{eqnarray} where we set $\hat{\z}_{i}^{l}=1(-1)$ if its probability predicted by Eq.~\ref{eq:Estep} is larger(no more) than $0.5$, $anno(\x_i)$ is the number of queried annotations of instance $\x_i$ and $\xi\in (0,1)$ is a small constant to avoid zero devisor. We use the number of queried annotation to tradeoff LCI to avoid querying the same instance, and giving less queried instance more chance to be queried which may contain more unknown information. $\xi\in (0,1)$ is set as $0.5$ in this paper.

 Given the selected instance $\x_\ast$  by Eq.~\ref{eq:instanceSel}, the label whose estimation is most uncertain is selected:
   \begin{eqnarray}\label{eq:labelSel}
   l^\ast = \arg\min_l CL(\x_{i^\ast},l) = |\; p( \z_{i^\ast}^{l}=1|{\x_{i^\ast}}) - 0.5\;|.
   \end{eqnarray} \textbf{Annotator Selection} Given the selected instance and label $(\x_\ast,l^\ast)$ by Eq.~\ref{eq:instanceSel} and Eq.~\ref{eq:labelSel}, we need to select the most reliable annotator on it to collect high quality annotations. Eq.~\ref{eq:expertise} provides the information about how reliable each annotator would be on each instance and label, which is inferred from observed annotations by our crowsourcing model. So we compute the expertise level of all annotators on the selected instance-label pair and select the most reliable one:
  \begin{eqnarray}\label{eq:annotatorSel}
  j^\ast = \arg\max_j CA(\x_{i^\ast},l^\ast, \w_{j}^{l^\ast}) = \e_{{i^\ast} j}^{l^\ast}= \sigma( (\w_{j}^{l^\ast})^{'}{\x_{i^\ast}}).  \end{eqnarray} After the instance-label pair and the best annotator on it are selected, the annotation is queried and added to the training data to update the multi-label crowdsourcing model. The MAC (Multi-label Active learning from Crowds) algorithm is summarized in Algorithm 1.

 \begin{algorithm}[t]
 \caption{The MAC Algorithm} \label{MAC}
 \begin{algorithmic}[1]
 \STATE \textbf{\textsc{Input}}: \\
 \STATE $\;\;\;\;$  training set $D = \{D_l, D_u\}$, parameters $\lambda$, $k$ \\
 \STATE \textbf{\textsc{Train}}:\\
 \STATE $\;\;\;\;$ \textbf{initialization}
 \STATE $\;\;\;$  get the enhanced representation $\hat{\x}= [\x,\c]$ for instance $\x \in D_l$ by Eq.~\ref{eq:labelCorCode} \\
 \STATE $\;\;\;$  get the initial estimation of $\{\w_{0}^{l},\w_{j}^{l}\}$ using LIBLINEAR on $D_l$\\
 \STATE $\;\;\;\;$  \textbf{repeat}:\\
 \STATE $\;\;\;\;\;\;\;\;$  get the label estimations for instances $\x$ in $D_u$ by Eq.~\ref{eq:Estep} with $\{\w_{0}^{l}\}$ and $ \{\w_{j}^{l}\}$ \\
  \STATE $\;\;\;\;\;\;\;\;$  select instance $\x_{i^\ast}$ by Eq.~\ref{eq:instanceSel}\\
  \STATE $\;\;\;\;\;\;\;\;$  select label $l^\ast$ for instance $\x_{i^\ast}$ by Eq.~\ref{eq:labelSel}\\
  \STATE $\;\;\;\;\;\;\;\;$  select annotator $j^\ast$ for instance $\x_{i^\ast}$ on label $l^\ast$ by Eq.~\ref{eq:annotatorSel}\\
 \STATE $\;\;\;\;\;\;\;\;$  get the annotation $\y_{{i^\ast}{j^\ast}}^{l^\ast}$ for $\x_{i^\ast}$ on label $l^\ast$ from annotator $j^\ast$\\
  \STATE $\;\;\;\;\;\;\;\;$  remove annotator $j^\ast$ for instance $\x_{i^\ast}$ on label $l^\ast$ \\
  \STATE $\;\;\;\;\;\;\;\;$  remove label $l^\ast$  from instance $\x_{i^\ast}$ if all annotators on $l^\ast$ are removed \\
  \STATE $\;\;\;\;\;\;\;\;$  remove instance $\x_{i^\ast}$ from $D_u$ if all labels of $\x_{i^\ast}$ are removed\\
  \STATE $\;\;\;\;\;\;\;\;$  add $\x_{i^\ast}$ to $D_l$ if $\x_{i^\ast}$ is not in $D_l$, add ${\y_{{i^\ast}{j^\ast}}}^{l^\ast}$ to $D_l$ for $\x_{i^\ast}$ on label $l^\ast$ \\
  \STATE $\;\;\;\;\;\;\;\;$  update $\{\w_{0}^{l},\w_{j}^{l}\}$  on $D_l$\\
  \STATE $\;\;\;\;$  \textbf{util} the maximum number of queries is reached or $D_u$ is empty.

  \STATE \textbf{\textsc{Test}}:\\
  \STATE $\;\;\;\;$  for instance $\x_t$, get its enhanced representation $\hat{\x_t}=[\x_t,\c_t]$ by Eq.~\ref{eq:labelCorCode} \\

 \STATE $\;\;\;\;$  get its prediction on label $l$ $\z_t^{l}$ by Eq.~\ref{eq:classifier}

 \end{algorithmic}
 \end{algorithm}\vspace{-1mm}

 \subsection{Computational Complexity}
  The computational complexity of MAC composes of three parts: 1) the initialization step, we use LIBLINEAR~\cite{liblinear} to get the initial estimation of classifier and annotator parameters which is fast and accurate; 2) the active selection part by Eq. 15-17, %Eq.~\ref{eq:instanceSel,eq:labelSel,eq:annotatorSel},
  which is linear in the number of instance, the number of labels and the number of annotators; 3) model updating, after the queried annotation is added to the training data, the crowdsourcing model is updated using EM. The parameters are initialized by results from last step which makes the EM converges fast, usually in less than $100$ iterations. The computational complexity of E-step in Eq.~\ref{eq:Estep} is linear in the number of involved instances and the number of involved annotations;  the computation complexity of M-step in Eq.~\ref{eq:Q} depends on the employed optimization approach. We exploit gradient ascent to solve the M-step, which needs to iteratively compute the gradient of parameters until convergence or maximum iteration number is reached. At each iteration, the computation complexity is linear in the number of involved annotators, the number of involved instance and the number of involved annotation.

 \section{Experiments}
  In this section, we compare our MAC approach with a number baseline methods on two natural scene classification datasets in which each picture can be categorized into one or more classes. The Image dataset~\cite{MultiLabelImage}\footnote{\emph{http://mulan.sourceforge.net/datasets.html}} contains $2000$ natural scene images and $5$ possible labels \{\emph{desert, mountains, sea, sunset, trees}\}, each image having on average $1.24\pm0.44$ labels. The Scene dataset~\cite{dataimage}\footnote { \emph{http://cse.seu.edu.cn/people/zhangml/Resources.htm}} contains $2407$ images and $6$ possible labels \{\emph{beach, field, foliage, mountain, sunset, urban}\}, each image having on average $1.07\pm0.26$ labels. The images belonging to more than one class in these two datasets comprise about $22\%$ of the whole data.

  The parameter $\lambda$ is set as $1e^{-3}$ and $k$ is set as $10$ in the experiments. As there exists no other method for the problem of multi-label active learning from crowds, we compare our MAC approach with the following 2 groups of 7 baselines.

 \textbf{Group1} exploits the label correlations of multi-label data, i.e., the multi-label classifier and annotators are learned respectively on enhanced and original instance representations: 1) MCR+RD: learn the classifier and annotators using our multi-label crowdsourcing model in Figure~\ref{fig:multigf}; randomly select (instance, label, annotator) for labeling; 2) MV+ACT: instead of learning by our multi-label crowdsourcing model, use majority voting on the collected annotations and train a logistic regression classier; select the (instance, label) by our active selection component in section 3.2, randomly select one annotator for labeling; 3) MV+RD: instead of learning by our multi-label crowdsourcing model, use majority voting on the available annotations and train a logistic regression classier; randomly select (instance, label, annotator) for labeling;

 \textbf{Group2} does not exploit the label correlations of multi-label data, i.e., both the multi-label classifier and annotators are learned on original instance representations: 4) SCR+ACT: learn the classifier and annotators using the probabilistic model in Figure~\ref{fig:singlegf} on each label independently; actively select (instance, label, annotator) by our active selection component in section 3.2; 5) SCR+RD: learn the classifier and annotators using the probabilistic model in Figure~\ref{fig:singlegf} on each label independently; randomly select (instance, label, annotator) for labeling; 6) SMV+ACT: use majority voting on the available annotations and train a logistic regression classier; select the (instance, label) by our active selection component in section 3.2, randomly select one annotator for labeling; 7) SMV+RD: use majority voting on the available annotations and train a logistic regression classier; randomly select (instance, label, annotator) for labeling.

  $3$ labelers with different expertise are simulated to generate the crowdsourcing annotations. For each dataset, on one specific label $l$, we proceed as follows: we first train a logistic regression model for this label on the whole dataset by LIBLINEAR~\cite{liblinear}, then according to the probability output of the logistic regression model on the dataset, we cluster the data into three subsets by k-means. After that, each of the $3$ simulated annotators $i$, $i=1, 2, 3$ performs as an expert on the $i$-th cluster and gives the groundtruth label for annotation; for the remaining data belonging to the other clusters, its probability of correctly annotating the groundtruth label is $75\%$ (the labels for these data are randomly switched with probability $25\%$). By such an annotation generation process, the annotations provided by crowds for specific label is dependent on the semantic of this label. Furthermore, on each label, the expertise of annotators are different on different instances.

  For each dataset, we randomly partition the data into three parts which compose $2.5\%$, $47.5\%$ and $50\%$  of the whole dataset to respectively construct the initial annotated labeled training data, the unlabeled training data and the test data. The random partition is repeated for five times on each dataset and the average performance is reported. At each active query, the (instance, label, annotator) triple is selected to get the annotation and then added into the annotated data. After every $200$ annotation queries the performance of the multi-label classifier on the test data is reported. The query process terminates after the number of queried annotations reaches $8,000$. The commonly used multi-label performance measure micro-F1~\cite{MultiLabelReviewZhou} is used in our experiments.

\begin{figure}[htbp]
  \centering
  \subfigure[Group1 Comparison]{
    \label{fig:imageadd1} %% label for first subfigure
    \includegraphics[width=1.85in,height=1.3in]{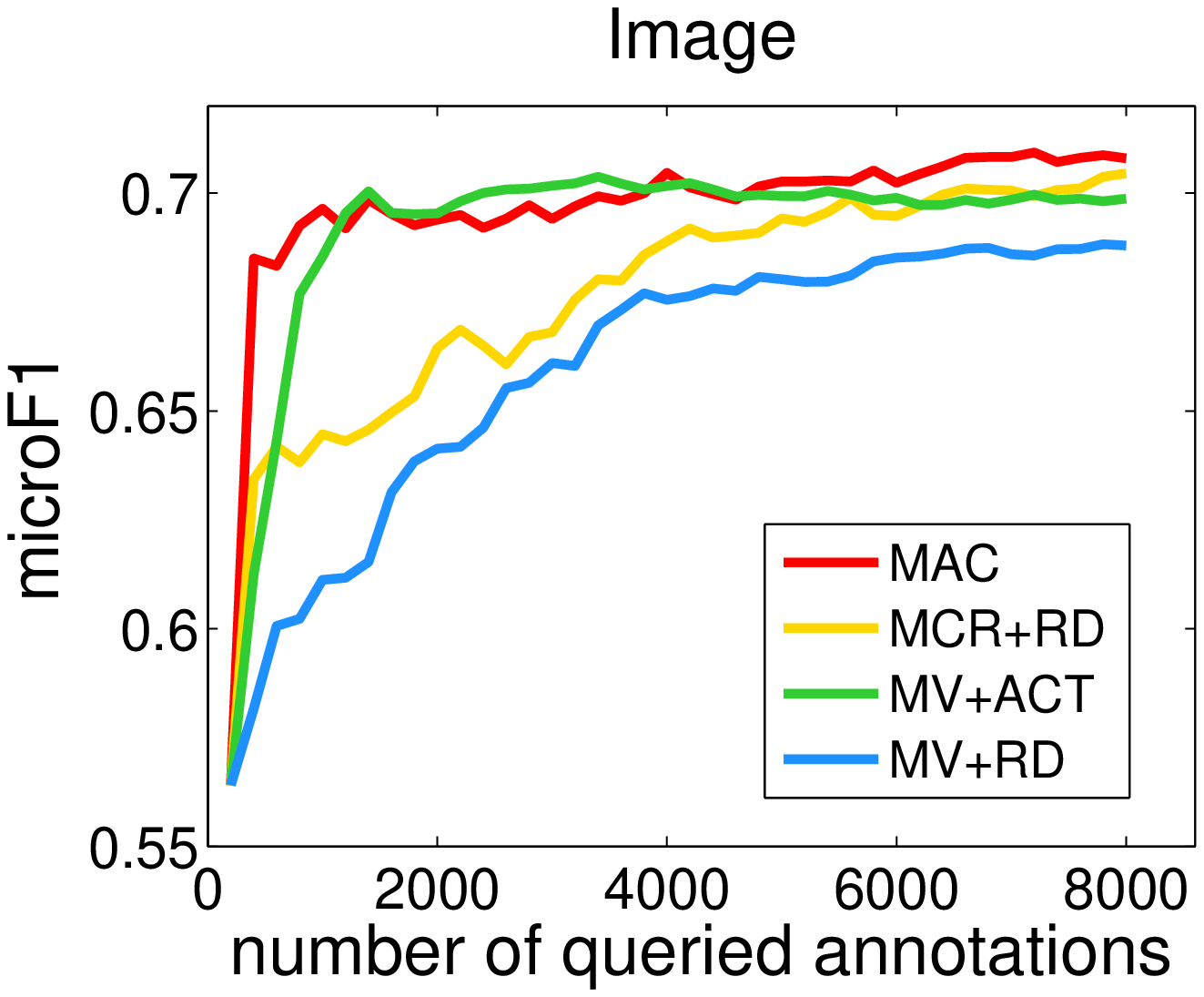} }
  \subfigure[Group2 Comparison]{
    \label{fig:imageadd0} %% label for second subfigure
   \includegraphics[width=1.85in,height=1.3in]{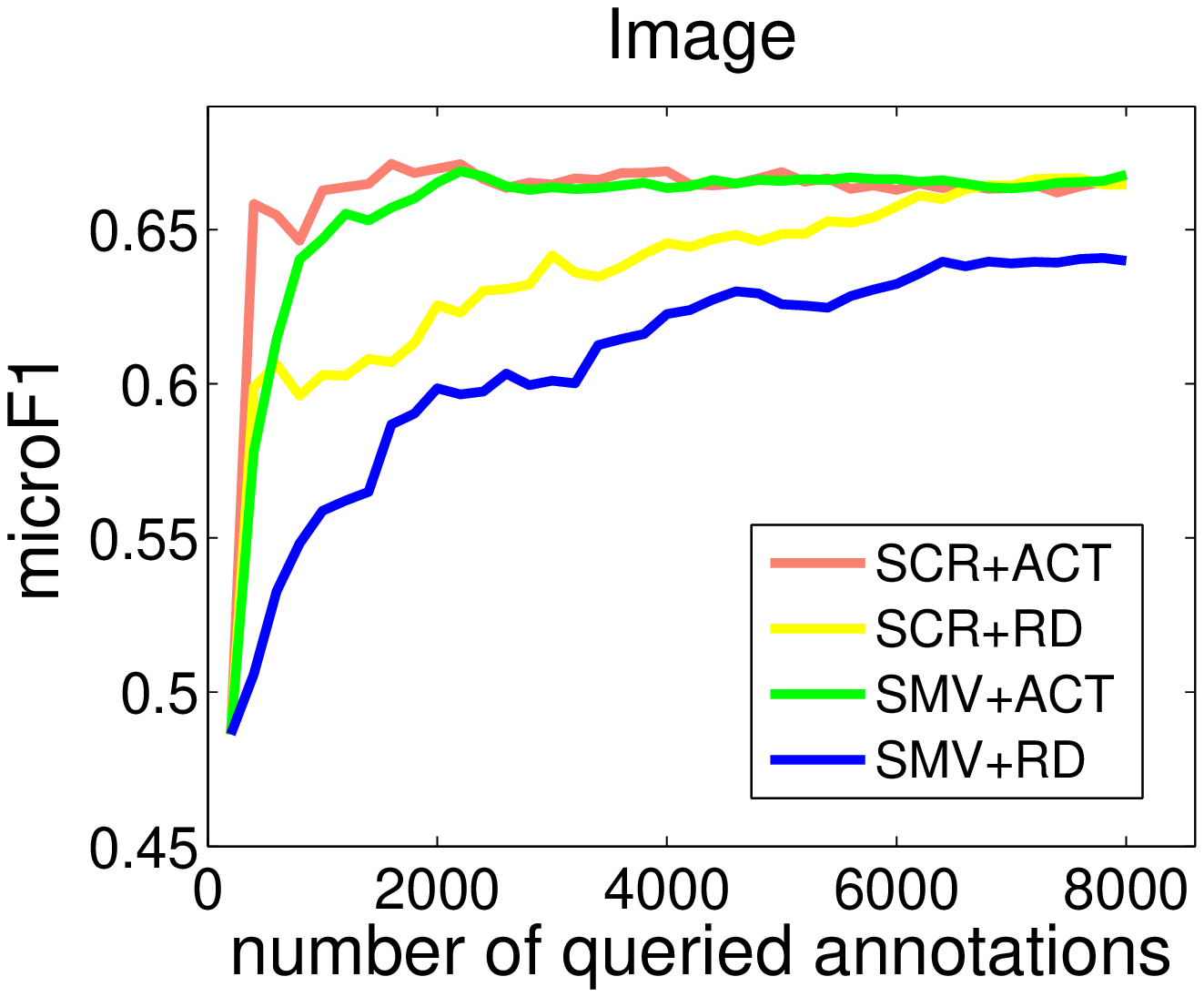}}
  \subfigure[Group1 v.s. Group2]{
    \label{fig:imageaddvs} %% label for second subfigure
  \includegraphics[width=1.85in,height=1.3in]{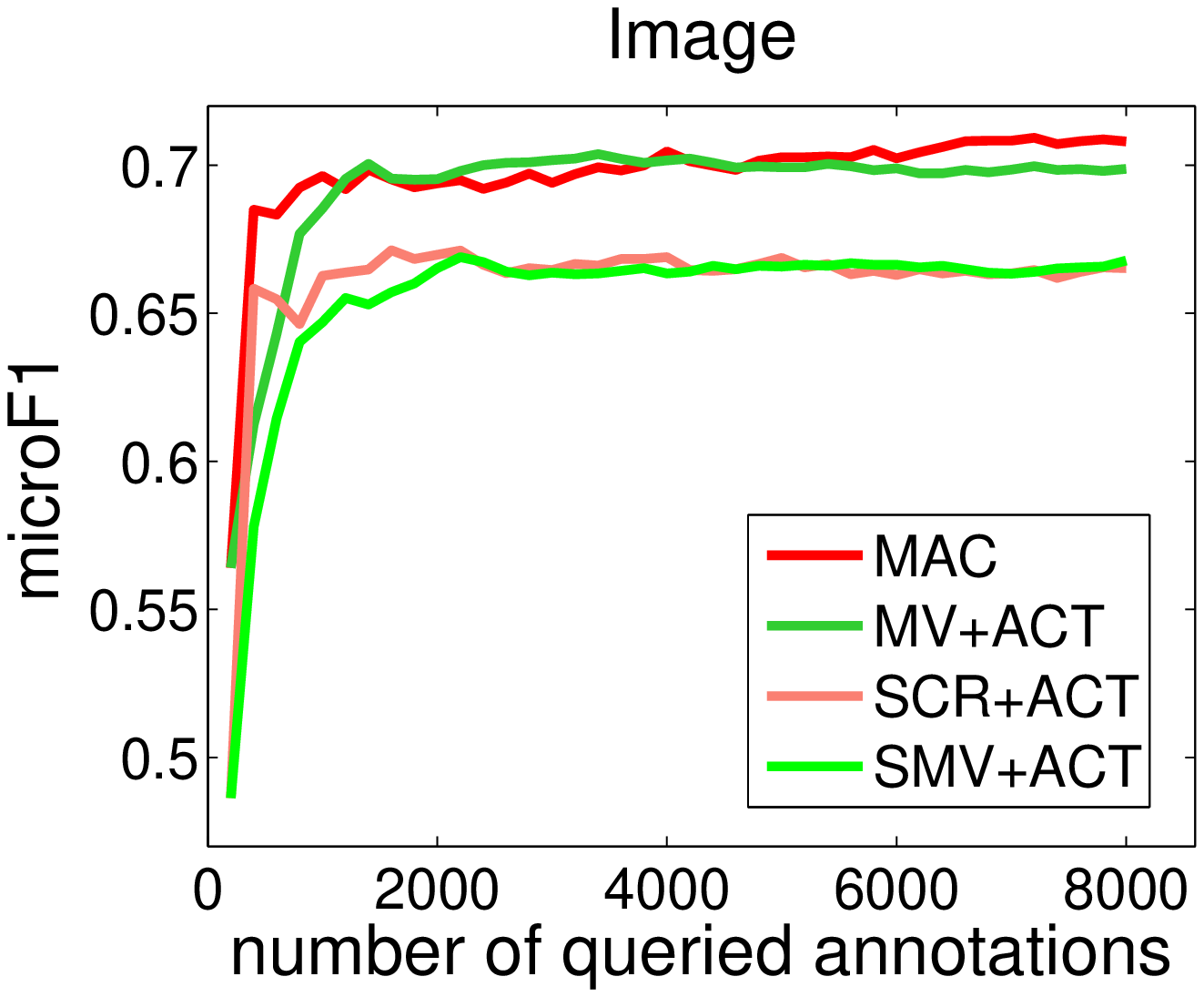}}
  \caption{The average results over 5 runs in terms of micro-F1(the larger the better) on Image.}
  \label{fig:imageresult} %% label for entire figure
\end{figure}

\begin{figure}[htbp]
  \centering
  \subfigure[Group1 Comparison]{
    \label{fig:sceneadd1} %% label for first subfigure
    \includegraphics[width=1.85in,height=1.3in]{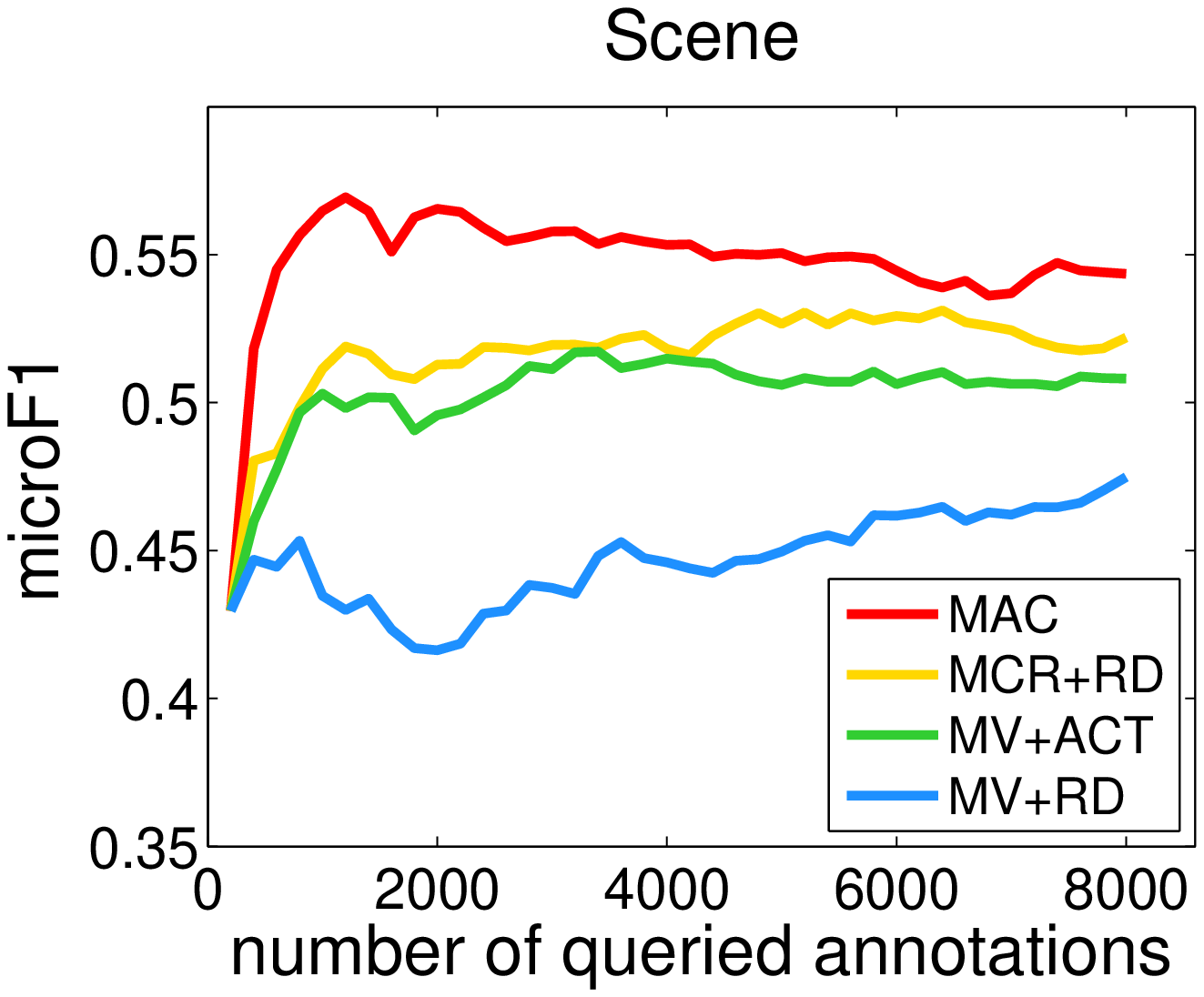} }
  \subfigure[Group2 Comparison]{
    \label{fig:sceneadd0} %% label for second subfigure
   \includegraphics[width=1.85in,height=1.3in]{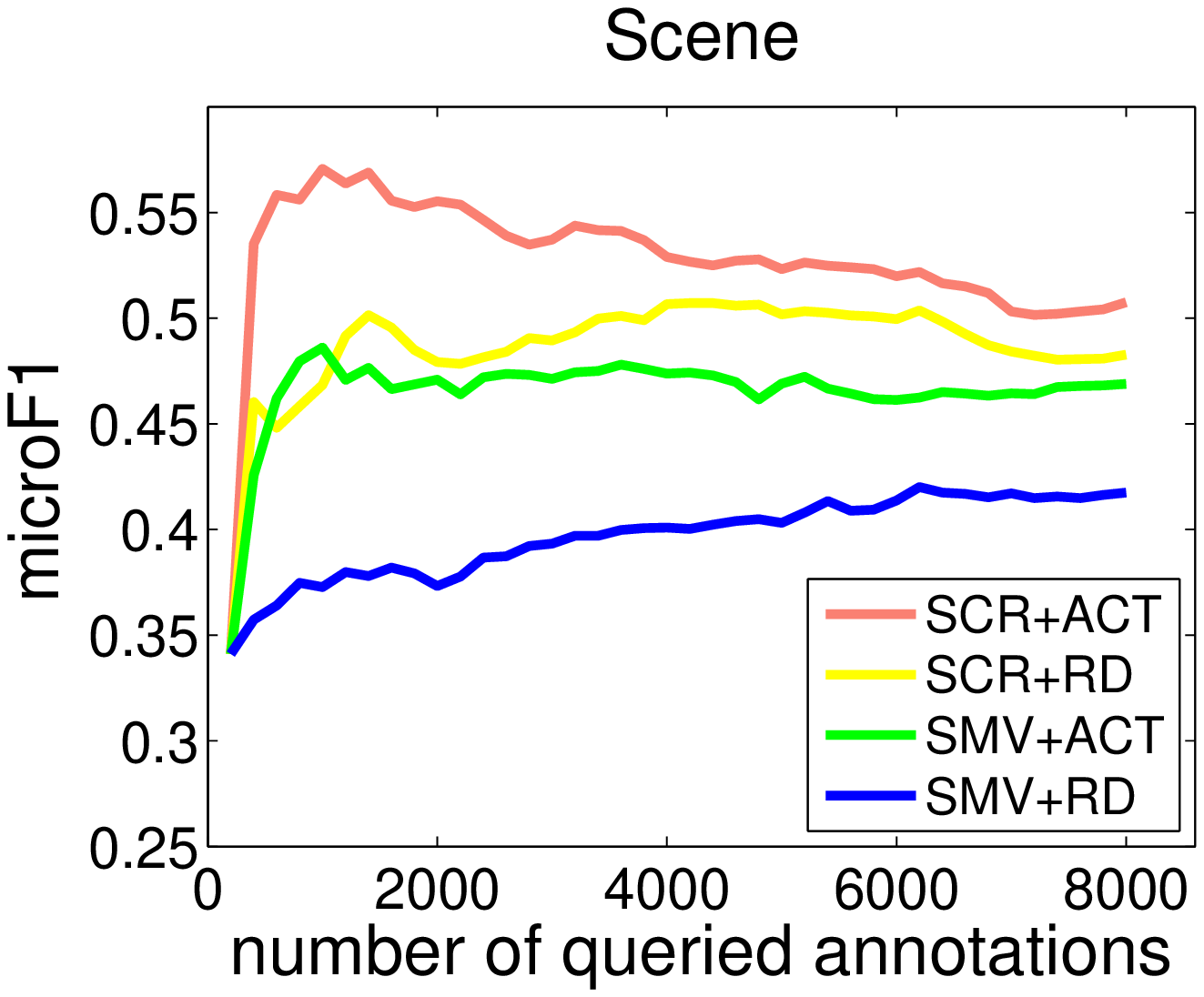}}
  \subfigure[Group1 v.s. Group2]{
    \label{fig:sceneaddvs} %% label for second subfigure
  \includegraphics[width=1.85in,height=1.3in]{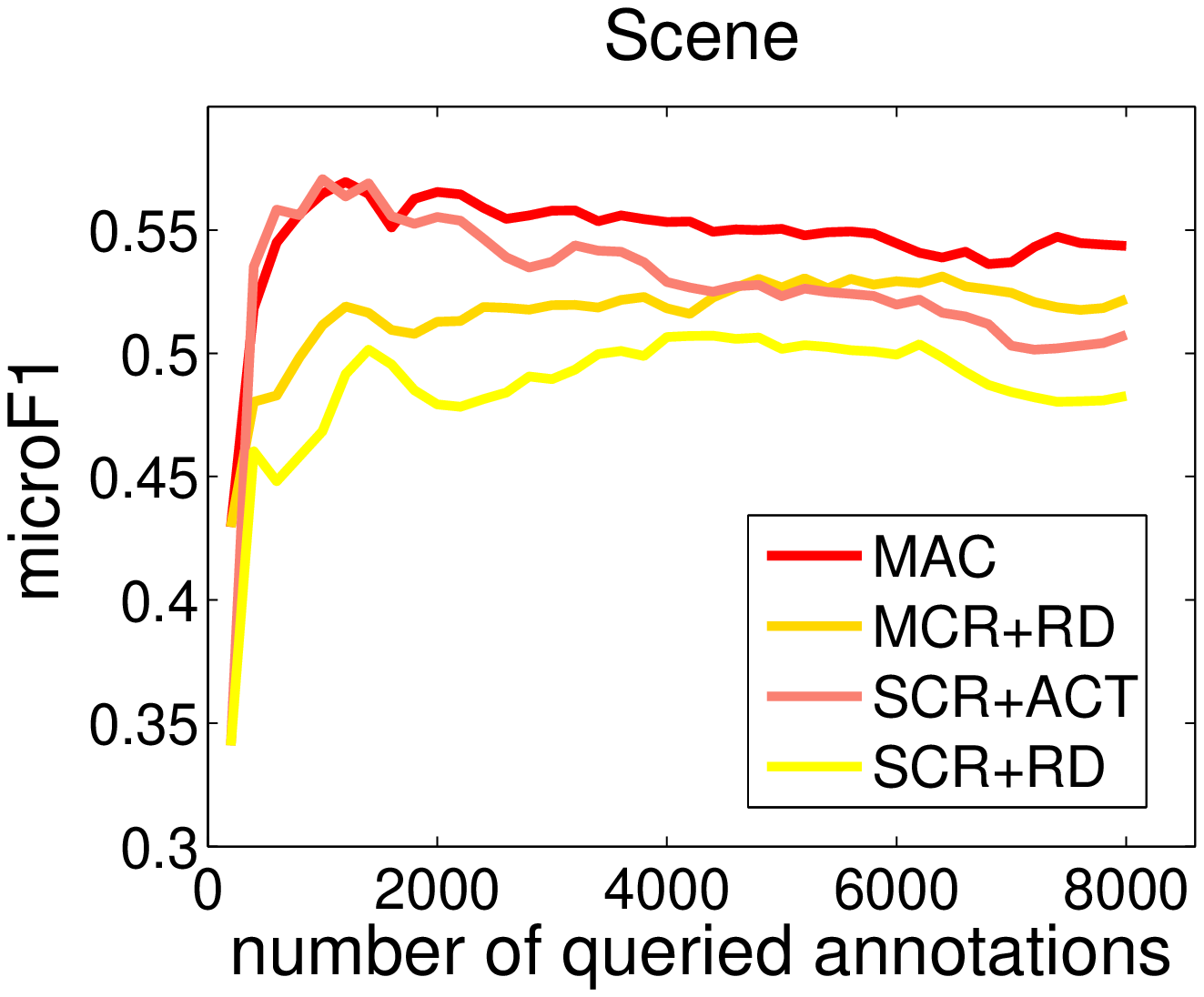}}
  \caption{The average results over 5 runs in terms of micro-F1(the larger the better) on Scene.}
  \label{fig:sceneresult} %% label for entire figure
\end{figure}

  In Figure~\ref{fig:imageadd1}/\ref{fig:sceneadd1} and Figure~\ref{fig:imageadd0}/\ref{fig:sceneadd0}, we test our crowdsourcing learning component and the active selection component between methods within Goup1 and Group2. In Figure~\ref{fig:imageaddvs}/\ref{fig:sceneaddvs} we test the label relationship role by comparing our MAC with the best baseline in Group1 and Group2.

  From Figure~\ref{fig:imageadd1} and Figure~\ref{fig:imageadd0}, we can see that: 1) compared to random sampling, our active selection component clearly helps improve learning for both our crowdsourcing learning model and majority voting (MAC vs MCR+RD; MV+ACT vs MV+RD; SCR+ACT vs SCR+RD; SMV+ACT vs SMV+RD); 2) compared to majority voting, our crowdsourcing model aggregates crowds better, especially when the annotations quality can not be guaranteed (MCR+RD vs MV+RD; SCR+RD vs SMV+RD). From Figure~\ref{fig:imageaddvs} in which MAC is compared with the best method in Group1 (MV+ACT) and Group2 (SCR+ACT, SMV+ACT), it can be seen that utilizing the label relationship for classifier learning greatly boost learning (MAC vs SCR+ACR; MV+ACT vs SMV+ACT), especially at early stage when the number of training data is small. Similar results for our MAC and baselines on Scene dataset are also obtained in Figure~\ref{fig:sceneresult}. It's not surprising that by utilizing the label relationship and active selection, our MAC approach achieves the best learning performance.


 \section{Conclusion}
  In this paper, we consider the problem of reducing the labeling cost of multi-label learning by actively learning from crowdsourcing annotations, where during the multi-label active query process, rather than resorting to a high cost oracle for the ground-truth, the labels are provided by multiple low-cost imperfect annotators. To deal with this problem, we propose the MAC (Multi-label Active learning from Crowds) approach which encodes the local influence of label correlations to build a probabilistic multi-label crowdsourcing model and then propose an active selection criterion to query the best annotator for the most valuable instance. Experimental results show that our approach is more effective than baselines. Currently, the computational complexity of MAC is linearly dependent on the number of labels; in the future, we plan to develop more efficient methods for large number of labels.

\bibliography{nips15}\bibliographystyle{alpha}

\newcommand{\etalchar}[1]{$^{#1}$}
\begin{thebibliography}{WRW{\etalchar{+}}09}

\bibitem[BLSB04]{MultiLabelImage}
M.~R. Boutell, J.~Luo, X.~Shen, and C.~M. Brown.
\newblock Learning multi-label scene classification.
\newblock {\em Pattern Recognition}, 37(9):1757--1771, 2004.

\bibitem[DLR77]{EM77}
A.~Dempster, N.~Laird, and D.~Rubin.
\newblock Maximum likelihood estimation from incomplete data.
\newblock {\em Journal of the Royal Statistical Society - B}, 39(1):1--38,
  1977.

\bibitem[ERH14]{CrowdcostAMV14}
Seyda Ertekin, Cynthia Rudin, and Haym Hirsh.
\newblock Approximating the crowd.
\newblock {\em Data Mining and Knowledge Discovery}, 28(5-6):1189--1221, 2014.

\bibitem[EW01]{MultiLabelGene}
A.~Elisseeff and J.~Weston.
\newblock A kernel method for multi-labelled classification.
\newblock In {\em Advances in Neural Information Processing Systems 14}, pages
  681--687, 2001.

\bibitem[FCH{\etalchar{+}}08]{liblinear}
Rong-En Fan, Kai-Wei Chang, Cho-Jui Hsieh, Xiang-Rui Wang, and Chih-Jen Lin.
\newblock Liblinear: A library for large linear classification.
\newblock {\em Journal of Machine Learning Research}, 9:1871--1874, 2008.

\bibitem[FYT14]{Crowdactivemoe14}
Meng Fang, Jie Yin, and Dacheng Tao.
\newblock Active learning for crowdsourcing using knowledge transfer.
\newblock In {\em Proceedings of the 28th AAAI Conference on Artificial
  Intelligence}, pages 1809--1815, 2014.

\bibitem[HL11]{MultLabelActiveacml11}
C.-W. Hung and H.-T. Lin.
\newblock Multi-label active learning with auxiliary learner.
\newblock In {\em Proceedings of the 3th Asian Conference on Machine Learning},
  pages 315--330, 2011.

\bibitem[HZ13]{MultLabelActiveicdm13}
S.-J. Huang and Z.-H. Zhou.
\newblock Active query driven by uncertainty and diversity for incremental
  multi-label learning.
\newblock In {\em Proceedings of the 13th IEEE International Conference on Data
  Mining}, pages 1079--1084, 2013.

\bibitem[KOS11]{KOS11}
D.~R. Karger, S.~Oh, , and D.~Shah.
\newblock Iterative learning for reliable crowdsourcing systems.
\newblock In {\em Advances in Neural Information Processing Systems 24}, pages
  1953--1961, 2011.

\bibitem[LG13]{MultLabelActiveijcai13}
X.~Li and Y.~Guo.
\newblock Active learning with multi-label svm classification.
\newblock In {\em Proceedings of the 23rd International Joint Conference on
  Artificial Intelligence}, pages 1479--1485, 2013.

\bibitem[McC99]{MultiLabelText}
A.~K. McCallum.
\newblock Multi-label text classification with a mixture model trained by {EM}.
\newblock In {\em Working Notes of the AAAI'99 Workshop on Text Learning},
  pages 17--26, 1999.

\bibitem[RYZ{\etalchar{+}}10]{Raykar10}
V.C. Raykar, S.~Yu, L.H. Zhao, G.H. Valadez, C.~Florin, L.~Bogoni, and L.~Moy.
\newblock Learning from crowds.
\newblock {\em Journal of Machine Learning Research}, 11:1297--1322, 2010.

\bibitem[SCC08]{MultLabelActiveSigh08}
M.~Singh, E.~Curran, and P.~Cunningham.
\newblock Active learning for multi-label image annotation.
\newblock In {\em Proceedings of the 19th Irish Conference on Artificial
  Intelligence and Cognitive Science}, 2008.

\bibitem[SOJN08]{CrowdSourcingSnowNLP08}
R.~Snow, B.~O¡¯Connor, D.~Jurafsky, and A.~Y. Ng.
\newblock Cheap and fast¡ªbut is it good? evaluating non-expert annotations for
  natural language tasks.
\newblock In {\em Proceedings of the Conference on Empirical Methods in Natural
  Language Processing}, pages 254--263, 2008.

\bibitem[WBBP10]{Welinder10}
P.~Welinder, S.~Branson, S.~Belongie, and P.~Perona.
\newblock The multidimensional wisdom of crowds.
\newblock In {\em Advances in Neural Information Processing Systems 23}, pages
  2024--2432, 2010.

\bibitem[WRW{\etalchar{+}}09]{Whitehill09}
J.~Whitehill, P.~Ruvolo, T.~Wu, J.~Bergsma, and J.~Movellan.
\newblock Whose vote should count more: Optimal integration of labels from
  labelers of unknown expertise.
\newblock In {\em Advances in Neural Information Processing Systems 22}, pages
  2035--2043, 2009.

\bibitem[YRFD11]{CrowdactiveYan11}
Y.~Yan, R.~Rosales, G.~Fung, and J.~Dy.
\newblock Active learning from crowds.
\newblock In {\em Proceedings of the 28th International Conference on Machine
  Learning}, pages 1161--1168, 2011.

\bibitem[ZLPM14]{DennyZhouICML14}
D.~Zhou, Q.~Liu, J.~C. Platt, and C.~Meek.
\newblock Aggregating ordinal labels from crowds by minimax conditional
  entropy.
\newblock In {\em Proceedings of the 31st International Conference on Machine
  Learning}, pages 262--270, 2014.

\bibitem[ZZ07]{dataimage}
M.-L. Zhang and Z.-H. Zhou.
\newblock Ml-knn: A lazy learning approach to multi-label learning.
\newblock {\em Pattern Recognition}, 40(7):2038--2048, 2007.

\bibitem[ZZ14]{MultiLabelReviewZhou}
M.-L. Zhang and Z.-H. Zhou.
\newblock A review on multi-label learning algorithms.
\newblock {\em IEEE Transactions on Knowledge and Data Engineering},
  26(8):1819--1837, 2014.

\end{thebibliography}
\end{document}